\documentclass{article}

     \PassOptionsToPackage{numbers, compress}{natbib}
\usepackage[final]{neurips_2023}

\usepackage[utf8]{inputenc} % allow utf-8 input
\usepackage[T1]{fontenc}    % use 8-bit T1 fonts
\usepackage{hyperref}       % hyperlinks
\usepackage{url}            % simple URL typesetting
\usepackage{booktabs}       % professional-quality tables
\usepackage{amsfonts}       % blackboard math symbols
\usepackage{nicefrac}       % compact symbols for 1/2, etc.
\usepackage{microtype}      % microtypography
\usepackage{xcolor}         % colors
\usepackage{graphicx}
\usepackage{comment}
\usepackage{subcaption}
\usepackage{multirow}
\usepackage{mathtools}
\hypersetup{
    colorlinks=true,
    linkcolor=cyan
    }

\usepackage{array}
\newcommand\Tstrut{\rule{0pt}{2.6ex}}         % = `top' strut
\newcommand\Bstrut{\rule[-0.9ex]{0pt}{0pt}}   % = `bottom' strut

\title{Analyzing Zero-Shot Abilities of Vision-Language Models on Video Understanding Tasks}

\author{%
  Avinash Madasu \\
  Intel Labs \\
  \texttt{avinash.madasu@intel.com} \\
  \And
  Anahita Bhiwandiwalla\\
  Intel Labs \\
  \texttt{anahita.bhiwandiwalla@intel.com} \\
  \And
  Vasudev Lal \\
  Intel Labs \\
  \texttt{vasudev.lal@intel.com} \\
}

\begin{document}

\maketitle

\begin{abstract}
 Foundational multimodal models pre-trained on large scale image-text pairs or video-text pairs or both have shown strong generalization abilities on downstream tasks. However unlike image-text models, pretraining video-text models is always not feasible due to the difficulty in collecting large-scale clean and aligned  data, and exponential computational costs involved in the pretraining phase. Therefore, the pertinent question to ask is: Can image-text models be adapted to video tasks and is there any benefit to using these models over pretraining directly on videos? In this work, we focus on this question by proposing a detailed study on the generalization abilities of image-text models when evaluated on video understanding tasks in a zero-shot setting. We investigate 9 foundational image-text models on a diverse set of video tasks that include video action recognition (video AR), video retrieval (video RT), video question answering (video QA), video multiple choice (video MC) and video captioning (video CP). Our experiments show that image-text models exhibit impressive performance on video AR, video RT and video MC. Furthermore, they perform moderately on video captioning and poorly on video QA. These findings shed a light on the benefits of adapting foundational image-text models to an array of video tasks while avoiding the costly pretraining step. The code is available at \url{https://github.com/IntelLabs/multimodal_cognitive_ai/tree/main/Video-Zeroshot}.
%These findings 
\end{abstract}

\begin{comment}
Recently, there has been a tremendous increase in the capabilities of Vision Models largely attributed to the availability of massive scale image-text data on which these models are trained. In comparison to image-text data, the availability and access to video data is much lesser. In this work, we explore the robustness and zero-shot performance of models trained on image-text pairs when evaluated on video tasks. We show that with a few simple modifications to the <TODO> the models trained on image-text pairs are capable of exhibiting impressive performance on video related tasks such as video retrieval, action recognition, video multiple choice question answering, video question and answering and video generative captioning tasks. The robustness of these models can serve as a precedent to expanding the video task capabilities of models not trained on explicitly on video data.
\end{comment}

\section{Introduction}
% - Models trained on large scale image-text learn richer features than limited video data
% - These models are shown to be effective on video retrievals. 
% - Use image features for other video tasks
% - Benchmark img-text trained models zeroshot performance on video tasks 
% - Start with simpler Action Recog, Video MCQ, Video Retrieval, VideoQA, generative captioning tasks

Foundational vision-language (VL) models ~\cite{radford2021learning, Yang_2022_CVPR, yuan2021florence, li2022align, li2022blip, li2021supervision} have achieved state-of-the-results on multi-modal tasks like image-retrieval ~\cite{young2014image, lin2014microsoft} image captioning ~\cite{agrawal2019nocaps, lin2014microsoft} and visual question answering ~\cite{antol2015vqa, das2017visual} etc. 
These models are pretrained on large amounts of image-text pairs using contrastive learning and then evaluated on these tasks in a zero-shot or few-shot setting. However, pre-training such large scale models is extremely hard for videos because: (i) Most of the current large scale video-text datasets ~\cite{miech2019howto100m} are very noisy and mis-aligned (i.e. video frames and captions don't match temporally). Therefore, its difficult to collect clean and aligned video-text data without human-in-the-loop unlike image-text data. (ii) Pre-training models on such video-text datasets require heavy computational resources. 
Hence given the remarkable transfer capabilities of contrastive image-text models, the pertinent question to explore is:

\textit{How beneficial are foundational image-text models when applied on video tasks in a zero-shot setting?}

\begin{figure}
    \centering
    \includegraphics[width=0.9\textwidth]{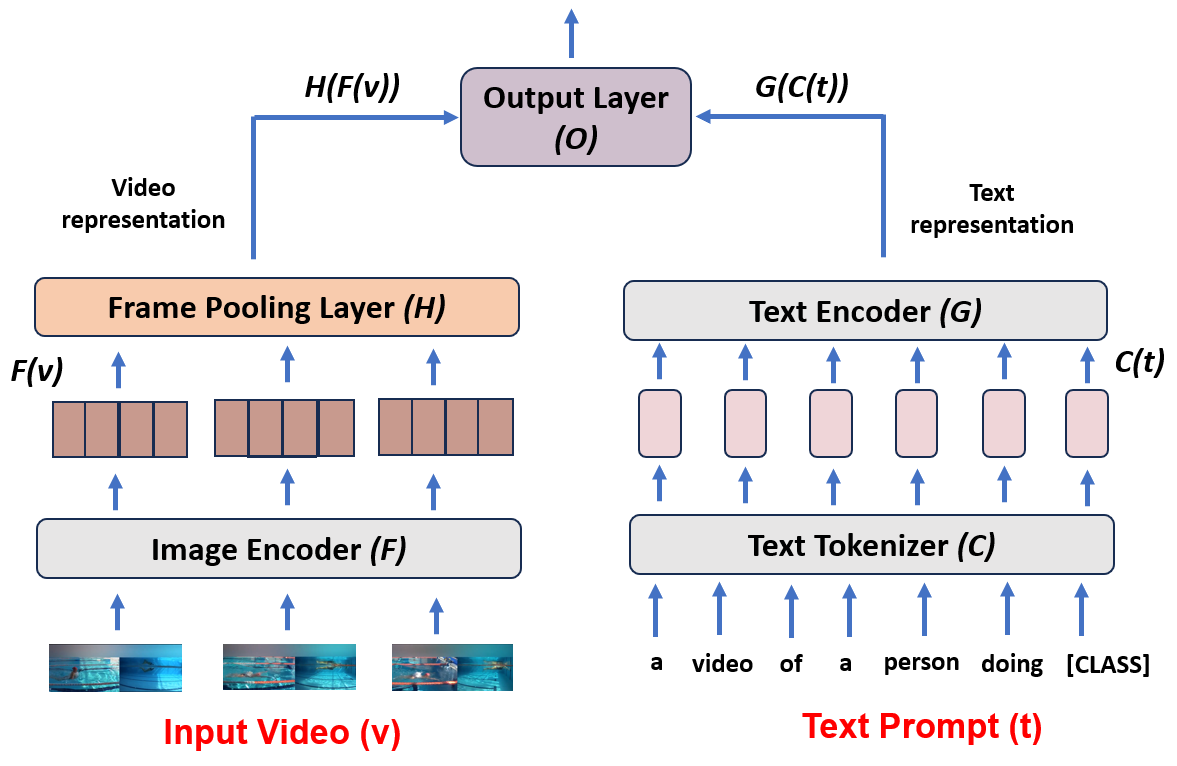}
    \caption{Overview of the architecture. It consists of four components (i) Image encoder \textit{(F)} to get frame representations for the input video, (ii) Text encoder \textit{(G)} to get text representations for the input text prompt, (iii) Frame pooling layer \textit{(H)} to obtain pooled frame representations, (iv) Final output layer \textit{(O)} to acquire the output.}
    \label{fig:arch}
\end{figure}

To investigate this, we conduct a systematic study by evaluating a diverse set of image-text models on a wide range of video tasks. 
We propose a simple architecture (Fig \ref{fig:arch}) to effectively adapt the image-text models for video-tasks without any changes to the architecture. For this study, we utilize 9 foundational image-text models and test these extensively on 5 video understanding tasks namely video action recognition (video AR), video retrieval (video RT), video question answering (video QA), video multiple choice (video MC) and video captioning (video CP) in a zero-shot setting. In addition, we also benchmark the zero-shot performance of image-text to the existing state-of-the-art video-text models on each of the tasks. Our zero-shot evaluation results demonstrate that:
\begin{enumerate}
    \item Foundational image-text models perform competitively to SotA video-text models on video action recognition, video retrieval and video multiple choice tasks.
    \item Image-text models struggle on complex video reasoning tasks such as video question answering while demonstrating reasonable performance on video captioning.
    \item The performance is heavily dependent on the size of the data used in pre-training image-text models and a frame count ranging from 12 to 20 is sufficient for the optimal score. 
\end{enumerate}

\section{Related Work}
\textbf{Vision-Language Models (VLM)}: There has been a body of work ~\cite{li2019visualbert, li2020hero, li2020oscar, li2021align,li2020unimo, dou2022empirical}that explore the phenomenon of pre-training on large image-text datasets ~\cite{sharma2018conceptual, changpinyo2021conceptual} and fine-tuning on downstream tasks. The goal is to learn generalized multi-modal representations applied to image understanding tasks ~\cite{deng2009imagenet, bossard2014food, fei2004learning, krizhevsky2009learning, antol2015vqa, agrawal2019nocaps, young2014image}. Unlike the previous works, we explore the generalization abilities of image-text models on a broad range of video understanding tasks in a zero-shot setting.

\textbf{Video understanding}: Several benchmarks have been introduced such as video action recognition ~\cite{soomro2012ucf101, kuehne2011hmdb, smaira2020short}, video retrieval ~\cite{xu2016msr, anne2017localizing, chen2011collecting}video question answering ~\cite{xu2017video, alamri2019audiovisual}, video captioning ~\cite{lei2020tvr, ZhXuCoAAAI18} etc. Transformer based models ~\cite{bain2021frozen, fu2021violet, li2023lavender} pre-trained on large scale video datasets ~\cite{bain2021frozen, miech2019howto100m} have been introduced and tested on these benchmarks. Our aim is to avoid the costly pre-training step and effectively adapt the foundational image-text models to videos.

\section{Methodology} \label{arch}
The primary question we aim to investigate in this work is: \textbf{How beneficial are image-text representations to downstream video tasks in a zero-shot setting?} To explore this, we propose a simple architecture that takes video and prompt text as inputs and returns the most similar class label or generates a text caption. As shown in the Figure \ref{fig:arch}, it consists of four components: image encoder \textit{(F)}, text encoder \textit{(G)}, frame pooling layer \textit{(H)} and final output layer \textit{(O)}. We leverage the image and text encoders of the respective models listed in Section 4.2. Additionally we add a frame pooling layer and also an output layer.

\textbf{Image Encoder \textit{(F)}.} Given a video $v$, we sample $N$ frames denoted as $v = [v_{1}, v_{2}, .., v_{N}]$. We then employ a pre-trained image encoder \textit{(F)} to get the representations for each of the sampled video frames $F(v) = [F(v_{1}), F(v_{2}), ... ,F(v_{N})]$. \\ \\
\textbf{Text Encoder \textit{(G)}.} Given a text prompt $t$, we use a pre-trained text encoder \textit{(G)} to obtain the text representations from the tokenized input text. \\ \\
\textbf{Frame Pooling Layer \textit{(H)}.} To obtain the final video representation, we leverage a frame pooling layer which computes the mean of all the frame representations.
\begin{equation}
    H(F(v)) =  \frac{\sum_{i=1}^{N} F(v_{i})}{N}
\end{equation}
\textbf{Output Layer \textit{(O)}.} The output layer takes the video and text representations and outputs a class label with the maximum cosine similarity score for classification tasks or generates a text for captioning tasks. 
\begin{comment}
\begin{equation}
    O_{class} = \arg \max_{k} \dfrac {H(F(v^{k})) \cdot G(C(t))} {\left\| H(F(v^{k}))\right\| _{2}\left\| G(C(t))\right\| _{2}}
\end{equation}
\begin{equation}
    O_{cap} = 
\end{equation}    
\end{comment}

\begin{table*}
\small
	\begin{center}
		%\resizebox{0.98\textwidth}{!}
		{
		\begin{tabular}{c c c c c c c} %
	    \hline
       \textbf{Task\Tstrut\Bstrut} & \textbf{Dataset\Tstrut\Bstrut} & \textbf{Split\Tstrut\Bstrut} & 
       \textbf{Test size\Tstrut\Bstrut} &
       \textbf{Metric\Tstrut\Bstrut} & \textbf{Frames\Tstrut\Bstrut} & \textbf{Classes\Tstrut\Bstrut} \\
        \hline
        \multirow{3}{*}{Action Recognition}
	    & Kinetics 700-2020\Tstrut ~\cite{smaira2020short} &  Val\Tstrut & 30850\Tstrut & Accuracy\Tstrut & 16\Tstrut & 700\Tstrut   \\
     & UCF101 ~\cite{soomro2012ucf101} &   Test & 13320 & Accuracy & 16 & 101  \\
     & HMDB51\Bstrut ~\cite{kuehne2011hmdb} &   Test\Bstrut & 3783\Bstrut & Accuracy\Bstrut & 16\Bstrut  & 51\Bstrut     \\
	\hline
    \multirow{3}{*}{Video Retrieval}
	    & MSRVTT\Tstrut ~\cite{xu2016msr} &  Test\Tstrut & 1000 \Tstrut & R@1 \Tstrut& 16\Tstrut  & 1000\Tstrut  \\
     & MSVD  ~\cite{chen2011collecting} &  Test & 670 & R@1 & 16 & 670  \\
     & DiDeMo\Bstrut ~\cite{anne2017localizing} &  Test\Bstrut & 1002\Bstrut & R@1\Bstrut & 32\Bstrut      & 1002\Bstrut \\
	\hline
 \multirow{2}{*}{Video QA}
	    & MSRVTT\Tstrut ~\cite{xu2017video}  &  Test \Tstrut & 67770\Tstrut & Accuracy\Tstrut & 16\Tstrut  & 1500\Tstrut  \\
     & MSVD\Bstrut ~\cite{xu2017video} &   Test\Bstrut & 11983\Bstrut & Accuracy\Bstrut & 16\Bstrut  & 1000\Bstrut \\
	\hline
 \multirow{4}{*}{Video MC}
	    & TGIF Transition\Tstrut ~\cite{jang2017tgif} & Test\Tstrut & 6232\Tstrut & Accuracy\Tstrut & 16\Tstrut & 5\Bstrut   \\
     & TGIF Action ~\cite{jang2017tgif} &  Test & 6232 & Accuracy & 16 & 5\Bstrut   \\
     & TGIF Frame ~\cite{jang2017tgif} &   Test & 6232 & Accuracy & 16 & 1540\Bstrut\\
     & MSRVTT\Bstrut ~\cite{yu2018joint} & Test\Bstrut & 2981\Bstrut & Accuracy\Bstrut & 16\Bstrut & 4\Bstrut\\
	\hline
 \multirow{2}{*}{Video Captioning}
	    & MSRVTT\Tstrut ~\cite{xu2016msr} &  Test\Tstrut & 2990\Tstrut & B-4, METEOR\Tstrut & 16\Tstrut  & Generation \Tstrut  \\
     & MSVD\Bstrut ~\cite{chen2011collecting} &     Test\Bstrut & 690\Bstrut & B-4, METEOR \Bstrut& 16\Bstrut & Generation \Bstrut  \\
	   \hline
		\end{tabular}}
		\caption{\textbf{Summary of the tasks and the datasets used in zero-shot evaluation.} The table also illustrates the dataset split, size of the test data, evaluation metric, number of video frames and number of classes for each of the datasets. R@1 denotes Recall-1, B-4 is the average of BLEU-1, BLEU-2, BLEU-3 and BLEU-4.}
        \label{tab:dataset_stats}
	\end{center}
\end{table*}

\begin{table*}
\small
	\begin{center}
		%\resizebox{0.98\textwidth}{!}
		{
		\begin{tabular}{c c c} %
	    \hline
       \textbf{Task\Tstrut\Bstrut} & \textbf{Dataset\Tstrut\Bstrut} & \textbf{Prompt\Tstrut\Bstrut} \\
        \hline
        \multirow{3}{*}{Action Recognition}
	    & Kinetics 700-2020\Tstrut &  \texttt{a video of a person doing \{CLASS NAME\}}\Tstrut    \\
     & UCF101  &  \texttt{a video of a person doing \{CLASS NAME\}}    \\
     & HMDB51\Bstrut &   \texttt{a demonstration of a person doing \{CLASS NAME\}}\Bstrut       \\
	\hline
    \multirow{3}{*}{Video Retrieval}
	    & MSRVTT\Tstrut &  \texttt{\{QUERY\}}\Tstrut    \\
     & MSVD  &   \texttt{\{QUERY\}}   \\
     & DiDeMo\Bstrut &   \texttt{\{QUERY\}}\Bstrut       \\
	\hline
 \multirow{2}{*}{Video QA}
	    & MSRVTT - QA\Tstrut &    \texttt{Question: \{QUESTION\}  Answer: \{ANSWER CANDIDATE\}} \Tstrut  \\
     & MSVD - QA \Bstrut &   \texttt{Question: \{QUESTION\}  Answer: \{ANSWER CANDIDATE\}}\Bstrut   \\
	\hline
 \multirow{3}{*}{Video MC}
	    & TGIF Transition\Tstrut &    \texttt{\{QUESTION\} \{ANSWER CANDIDATE\}} \Tstrut \\
     & TGIF Action  &   \texttt{\{QUESTION\} \{ANSWER CANDIDATE\}}  \\
     & TGIF Frame  &   \texttt{\{QUESTION\} \{ANSWER CANDIDATE\}}  \\
     & MSRVTT - MC \Bstrut &      \texttt{\{QUESTION\} \{ANSWER CANDIDATE\}} \Bstrut  \\
	\hline
 \multirow{2}{*}{Video Captioning}
	    & MSRVTT \Tstrut &    \texttt{A video of}\Tstrut \\
     & MSVD \Bstrut  &    \texttt{A video of}\Bstrut \\
	   \hline
		\end{tabular}}
		\caption{\textbf{Summary of the tasks and prompts used in zero-shot evaluation.} The curly brackets in a prompt is replaced with all the class names for action recognition and answer candidates for Video QA and Video MC tasks.}
        \label{tab: data_prompt}
	\end{center}
\end{table*}

\section{Experimental Settings}
\subsection{Tasks}
We analyze five different video understanding tasks: video action recognition (video AR), video retrieval (video RT), video question answering (video QA), video multiple choice (video MC) and video captioning (video CP). Table ~\ref{tab:dataset_stats} summarizes the tasks, datasets for each task category, dataset split, metrics used for evaluation, number of frames and classes for each dataset. 
Below, we list the evaluation datasets for each of the tasks.\\\\
\textbf{Video Action Recognition (Video AR):} Kinetics 700-2020 ~\cite{smaira2020short}, UCF-101 ~\cite{soomro2012ucf101} and HMDB-51 ~\cite{kuehne2011hmdb}. \\
\textbf{Video Retrieval (Video RT):} MSRVTT ~\cite{xu2016msr}, MSVD ~\cite{chen2011collecting} and DiDeMo ~\cite{anne2017localizing}. \\
\textbf{Video Question Answering (Video QA):} MSRVTT-QA ~\cite{xu2017video} and MSVD-QA ~\cite{xu2017video}. \\
\textbf{Video Multiple Choice (Video MC):} TGIF-Transition, TGIF-Action, TGIF-Frame ~\cite{jang2017tgif} and MSRVTT-MC ~\cite{yu2018joint}. \\
\textbf{Video Captioning (Video CP):} MSRVTT ~\cite{xu2016msr} and MSVD ~\cite{chen2011collecting}.

\subsection{Models}
For experiments we use the following nine foundational image-text models: ALIP ~\cite{yang2023alip}, CLIP ~\cite{radford2021learning}, OpenCLIP ~\cite{ilharco_gabriel_2021_5143773}, SLIP ~\cite{mu2022slip}, LaCLIP ~\cite{fan2023improving}, BLIP-2 ~\cite{li2023blip}, InstructBLIP ~\cite{instructblip}, OFA ~\cite{wang2022ofa} and Pix2Struct ~\cite{lee2023pix2struct}.

\subsection{Evaluation Metrics}
For the tasks of video action recognition, video question answering and video multiple choice we use accuracy as the evaluation metric. We report the results using the standard R@1 (Recall@1) metric in case of video retrieval. We employ multiple metrics: B-4 ~\cite{papineni2002bleu}, METEOR ~\cite{banerjee2005meteor} and CIDEr ~\cite{vedantam2015cider} for the video captioning task. B-4 is the average of BLEU-1, BLEU-2, BLEU-3 and BLEU-4. 

\subsection{Implementation}

Table ~\ref{tab: data_prompt} summarizes the text prompts used for each of the video tasks.

\textbf{Video Action Recognition:} For the action recognition task, we use the prompts ``\texttt{a video of a person doing \{CLASS NAME\}}'' and  ``\texttt{a demonstration of a person doing \{CLASS NAME\}}''. The field \texttt{\{CLASS NAME\}} is substituted with all the class names present in the dataset and the cosine similarity is measured for the input video and all the classes. The predicted class is the one with maximum cosine similarity among all the candidate classes.

\textbf{Video Retrieval:} In case of video retrieval, the input query is used as the prompt and its similarity is calculated with all the videos in the database.

\textbf{Video QA and Video MC:} Regarding the tasks of video QA and video MC, the fields \texttt{\{Question\}} and \texttt{\{ANSWER CANDIDATE\}} in the prompt ``\texttt{Question: \{QUESTION\}  Answer: \{ANSWER CANDIDATE\}}'' are replaced with input question and all the possible answers respectively. The answer candidate for which the similarity between the input video and the text prompt is maximum is chosen as the predicted answer. 

\textbf{Video Captioning:} As for video captioning, the prompt ``\texttt{A video of}'' along with the video are used as inputs for generating captions.

\begin{table*}[t]
\scriptsize
	\begin{center}
		%\resizebox{0.98\textwidth}{!}
		{
		\begin{tabular}{c c c| c c c| c c c| c c} %
	    \hline
     & & & \multicolumn{3}{c}{\textbf{Video AR}}\Bstrut\Tstrut & \multicolumn{3}{c}{\textbf{Video Retrieval}}\Bstrut\Tstrut & \multicolumn{2}{c}{\textbf{Video QA}}\Bstrut\Tstrut \\
     \hline
         \textbf{Model\Tstrut\Bstrut} &
         \textbf{Type\Tstrut\Bstrut} &
         \textbf{\#PT\Bstrut\Tstrut} &
         \textbf{K700\Tstrut\Bstrut} & \textbf{UCF-101\Tstrut\Bstrut} & 
         \textbf{HMDB-51\Tstrut\Bstrut} &
         \textbf{MSRVTT\Tstrut\Bstrut} & \textbf{MSVD\Tstrut\Bstrut} & \textbf{DiDeMo\Tstrut\Bstrut} &
         \textbf{MSRVTT\Tstrut\Bstrut} &
         \textbf{MSVD\Tstrut\Bstrut}\\
        \hline
 ALIP\Tstrut\Bstrut ~\cite{yang2023alip} & IT\Tstrut\Bstrut & 15M\Bstrut\Tstrut & 15.7\Bstrut\Tstrut & 30.3\Bstrut\Tstrut & 21.2 \Bstrut\Tstrut & 15.3\Tstrut\Bstrut & 33.9\Tstrut\Bstrut & 14.3\Tstrut\Bstrut & 0.1\Tstrut\Bstrut & \textbf{7.7} \Tstrut\Bstrut  \\

 CLIP ~\cite{radford2021learning} & IT & 400M & \textbf{40.5} & \textbf{63.6} & \textbf{46.2} & 29.9 & 44.2&  22.6 & 1.5 & 6.5 \\

 OpenCLIP ~\cite{ilharco_gabriel_2021_5143773} & IT & 2B & 37.4 & 56.4 & 38.1 & 31.7& \textbf{48.2} & 22.6  & \textbf{2.9} & \textbf{7.7} \\

 SLIP ~\cite{mu2022slip} & IT & 15M & 13 & 15.2 & 19.8 & 14.1 & 30.3 &  17.2 & 1.5 & 4.1  \\
 LaCLIP\Bstrut ~\cite{fan2023improving} & IT\Bstrut & 400M\Bstrut & 39.1\Bstrut & 59.5 & 38.8 & \textbf{32} & 43.6\Bstrut &  \textbf{23.3} \Bstrut & 0.1\Bstrut & 5.5\Bstrut \\
 \hline
 VIOLET\Bstrut\Tstrut ~\cite{fu2023empirical} & VT \Bstrut\Tstrut  & 186M\Bstrut\Tstrut & - & - & - & 34.5\Bstrut\Tstrut & 48.3\Tstrut\Bstrut & 32.6\Tstrut\Bstrut & - & - \\
 LAVENDER ~\cite{li2023lavender} & VT  & 30M & - & -  & -  & - & -  & - & 2.7 & 9.2 \\
 MAXI\Bstrut ~\cite{lin2023match} & VT\Bstrut  & GPT3 & -\Bstrut & 78.2\Bstrut & 52.3\Bstrut & -\Bstrut &  -\Bstrut &  -\Bstrut &  -\Bstrut &  -\Bstrut\\
 
 \hline
 $\Delta$ & - & - & - & \textcolor{red}{-14.6} \Tstrut\Bstrut & \textcolor{red}{-5.9} \Tstrut\Bstrut & \textcolor{red}{-2.8} \Tstrut\Bstrut & \textcolor{red}{-0.1} \Tstrut\Bstrut & \textcolor{red}{-9.3} \Tstrut\Bstrut & \textcolor{green}{+0.2} \Tstrut\Bstrut & \textcolor{red}{-1.5} \Tstrut\Bstrut \\
 \hline
		\end{tabular}}
		\caption{\textbf{Results of zero-shot video action recognition, video retrieval and video question answering.} \textbf{Type} denotes whether the model is pre-trained on image-text (IT) or video-text (VT) data. \textbf{\#PT} implies the size of data used in pre-training. K700 is the kinetics 700-2020 ~\cite{smaira2020short} dataset. \textbf{$\Delta$} is the difference between the highest performance of image-text models and state-of-the-art video-text model.}
        \label{tab:act-ret-qa}
	\end{center}
\end{table*}

\begin{table*}[t]
\footnotesize
	\begin{center}
		%\resizebox{0.98\textwidth}{!}
		{
		\begin{tabular}{c c c c c c} %
	    \hline
         \textbf{Model\Tstrut\Bstrut} & \textbf{Type\Tstrut\Bstrut} & \textbf{TGIF-Transition\Tstrut\Bstrut} & \textbf{TGIF-Action\Tstrut\Bstrut} & \textbf{TGIF-Frame\Tstrut\Bstrut} & \textbf{MSRVTT-MC\Tstrut\Bstrut}\\
        \hline
ALIP\Tstrut ~\cite{yang2023alip} & Image-text \Tstrut  & 39.1\Tstrut & 38.5\Tstrut & 0.1\Tstrut & 62.4\Tstrut   \\
 CLIP ~\cite{radford2021learning} & Image-text & \textbf{51.3} & \textbf{50.5} & \textbf{5.8} & 74.3  \\
 OpenCLIP ~\cite{ilharco_gabriel_2021_5143773} & Image-text & 41.2 & 45.1 & 4.2  & 73.3 \\
 SLIP ~\cite{mu2022slip} & Image-text & 37.2 & 35.9 & 2.0   & 61.9  \\
  LaCLIP ~\cite{fan2023improving} \Bstrut & Image-text \Bstrut & 41.6 \Bstrut & 46.1 \Bstrut & 2.6 \Bstrut & \textbf{74.4} \Bstrut  \\
  \hline
  LAVENDER\Bstrut\Tstrut ~\cite{li2023lavender} & Video-text\Bstrut\Tstrut & 53.8\Bstrut\Tstrut & 55.1\Bstrut & 19.6\Bstrut\Tstrut & 87.2\Bstrut\Tstrut \\
 \hline
 $\Delta$ \Bstrut\Tstrut & - & \textcolor{red}{-2.5} \Bstrut\Tstrut & \textcolor{red}{-4.6} \Bstrut\Tstrut & \textcolor{red}{-13.8}\Bstrut\Tstrut & \textcolor{red}{-12.8} \\
 \hline
		\end{tabular}}
		\caption{\textbf{Results of zero-shot evaluation on video MC.} $\Delta$ is the difference between the highest performance of image-text models and state-of-the-art video-text model.}
        \label{tab: video_mul_choice}
	\end{center}
\end{table*}

\begin{table*}[t]
\footnotesize
	\begin{center}
		%\resizebox{0.98\textwidth}{!}
		{
		\begin{tabular}{c c c| ccc| ccc} %
	    \hline
      & & &  \multicolumn{3}{c}{\textbf{MSRVTT\Tstrut\Bstrut}} &\multicolumn{3}{c}{\textbf{MSVD\Tstrut\Bstrut}} \\
        \hline
	    \textbf{Model\Tstrut\Bstrut} & \textbf{Type\Tstrut\Bstrut} & \textbf{\#PT\Tstrut\Bstrut} & \textbf{B-4\Tstrut\Bstrut}& \textbf{METEOR\Tstrut\Bstrut}  & 
     \textbf{CIDER\Tstrut\Bstrut} & \textbf{B-4\Tstrut\Bstrut} & \textbf{METEOR\Tstrut\Bstrut} &\textbf{CIDER\Tstrut\Bstrut} \\
        \hline
        BLIP2\Tstrut ~\cite{li2023blip} & Image-text \Bstrut & 129M\Bstrut & \textbf{35} \Tstrut & \textbf{18} \Tstrut & \textbf{19} \Tstrut & \textbf{35.6}\Tstrut & \textbf{22.2}\Tstrut & \textbf{27.4}\Tstrut \\
        InstructBLIP ~\cite{instructblip}  & Image-text & 129.5M & 26.2 & 12.9 & 11.8 & 27.1 & 16.2 & 17.8\\
        OFA ~\cite{wang2022ofa} & Image-text & 36.27M & 18.7 & 10.5 & 4.8\Bstrut & 17.3 & 11.8 & -\Bstrut \\
        Pix2Struct ~\cite{lee2023pix2struct}\Bstrut & Image-text\Bstrut & 80M\Bstrut & 23.3\Bstrut & 14.4\Bstrut & -\Bstrut & 19.3\Bstrut & 15.3\Bstrut & -\Bstrut \\
	   \hline
    Vid2Seq \Tstrut\Bstrut ~\cite{yang2023vid2seq}  & Video-text \Tstrut\Bstrut & 1B + FT\Tstrut\Bstrut & - \Tstrut\Bstrut & 30.8 \Tstrut\Bstrut & 64.6 \Bstrut\Tstrut & - \Bstrut\Tstrut & 45.3 \Bstrut\Tstrut & 146.2 \Bstrut\Tstrut \\
    \hline
    $\Delta$ \Tstrut\Bstrut  & - \Tstrut\Bstrut & -\Tstrut\Bstrut & - \Tstrut\Bstrut & \textcolor{red}{-12.8} \Tstrut\Bstrut & \textcolor{red}{-45.6} \Bstrut\Tstrut & - \Bstrut\Tstrut & \textcolor{red}{-23.2} \Bstrut\Tstrut & \textcolor{red}{-118.8} \Bstrut\Tstrut \\
    \hline
		\end{tabular}}
		\caption{\textbf{Results of zero-shot evaluation on video captioning.} Note that Vid2Seq model is pretrained on YT-Temporal-1B ~\cite{zellers2022merlot} and then fine-tuned on MSRVTT and MSVD datasets respectively. $\Delta$ is the difference between the highest performance of image-text models and state-of-the-art video-text model.}
        \label{tab:video_cap}
	\end{center}
\end{table*}

\section{Results and Discussion}
\subsection{Zero-shot image-text models are competitive to SOTA video-text models on video AR, RT and MC tasks}
In Table ~\ref{tab:act-ret-qa}, we present the results of image-text models on video action recognition and video retrieval. Table ~\ref{tab: video_mul_choice} illustrates the results on video multiple choice. From the tables, we observe that image-text models perform competitively to state-of-the-art video-text models on most of the tasks. On the task of video action recognition, image-text models lag behind SotA video-text model by 14.6\% for UCF-101 and 5.9\% for HMDB-51 dataset. Note that, MAXI ~\cite{lin2023match} is pretrained on significantly larger video datasets compared to image-text models and is explicitly modelled to learn actions in them. 
%% Add a sentence how image-text models are not trained explicitly for action recognition

For the task of video retrieval, image-text models achieves comparable results to SotA video-text model. In fact, image-text models trail only by 2.8\% and 0.1\% for MSRVTT and MSVD respectively. In-case of DiDeMo, the difference in performance (9.2\%) is quite significant. Observe that, MSRVTT and MSVD are shorter length video datasets whereas DiDeMo is a paragraph to video retrieval dataset. 

In the case of video MC, we see that image-text models under perform by just 2.5\% and 4.6\% in case of TGIF-transition and TGIF-action datasets respectively. TGIF is a GIF dataset and are typically easier compared to videos and in addition the number of choices to select from in these datasets are limited to just 5. Hence, the image-text models were able to achieve similar performance to SotA video-text models on these relatively easier datasets. In contrast even though TGIF-frame is a GIF dataset, the frequency of choices is 1000 which are significantly higher. Therefore, we spot a notable difference (13.8\%) in performance between image-text and video-text models. We would like to add that the video-text models also perform considerably worse on TGIF-frame compared to TGIF-transition and TGIF-action. On MSRVTT-MC despite the limited multiple choices (i.e. 4), SotA video-text models evidently outperforms image-text models by 12.8\%. 
% Add this may be because of the difference in visual data (GIF vs video)

\subsection{Zero shot image-text models perform reasonably on video captioning.}
In Table \ref{tab:video_cap}, we provide a comparison between zero-shot image-text models and fine-tuned video-text models on the task of video captioning. As seen in the table, it is clear that zero-shot image-text models (BLIP-2) perform reasonably to video-text models despite pre-training on just 13\% data and not fine-tuning on respective datasets. 
\subsection{Image-text and video-text models perform worse on video QA in a zero-shot setting}
Table ~\ref{tab:act-ret-qa} illustrates the zero-shot results of image-text and video-text models on the video QA task. From the table, it is clear that video-text models marginally outperform image-text models in zero-shot video QA. It is worth noting that both image-text and video-text perform poorly on QA datasets. Video QA is a complex task which requires modelling of object and attribute relationships in the videos. This could be the reason zero-shot models show subpar performance.

\begin{figure}
     \centering
     \begin{subfigure}[b]{0.45\textwidth}
         \centering
         \includegraphics[width=\textwidth]{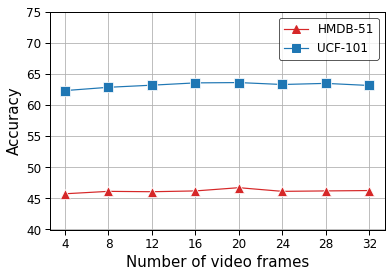}
         \caption{Video Action Recognition}
         \label{fig:clip-video-act-rec}
     \end{subfigure}
     \hfill
     \begin{subfigure}[b]{0.45\textwidth}
         \centering
         \includegraphics[width=\textwidth]{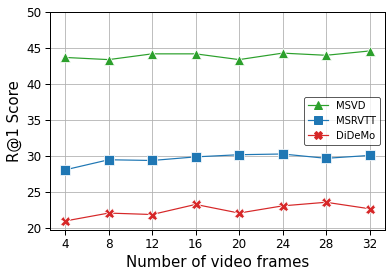}
         \caption{Video Retrieval}
         \label{fig:clip-video-ret}
     \end{subfigure}
        \caption{Figure shows the effect of number of video frames on the zero-shot performance of CLIP for video action recognition and video retrieval. Figures for the additional tasks and image-text models are included in the appendix.}
        \label{fig:clip_num_frames}
\end{figure}

\begin{figure}
     \centering
     \begin{subfigure}[b]{0.45\textwidth}
         \centering
         \includegraphics[width=\textwidth]{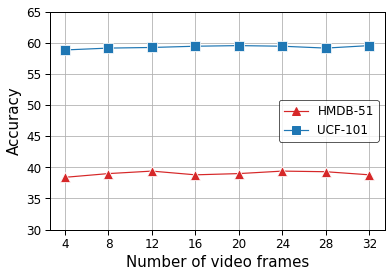}
         \caption{Video Action Recognition}
         \label{fig:laclip-video-act-rec}
     \end{subfigure}
     \hfill
     \begin{subfigure}[b]{0.45\textwidth}
         \centering
         \includegraphics[width=\textwidth]{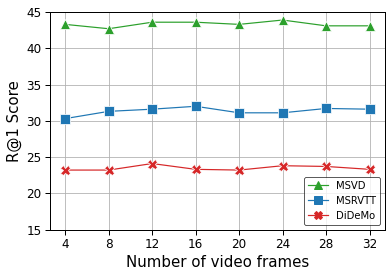}
         \caption{Video Retrieval}
         \label{fig:laclip-video-ret}
     \end{subfigure}
        \caption{Figure shows the effect of number of video frames on the zero-shot performance of LaCLIP for video action recognition and video retrieval.}
        \label{fig:laclip_num_frames}
\end{figure}

\begin{figure}
     \centering
     \begin{subfigure}[b]{0.49\textwidth}
         \centering
         \includegraphics[width=\textwidth]{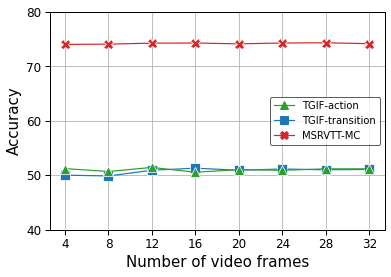}
         \caption{CLIP}
         \label{fig:clip-videomc}
     \end{subfigure}
     \hfill
     \begin{subfigure}[b]{0.49\textwidth}
         \centering
         \includegraphics[width=\textwidth]{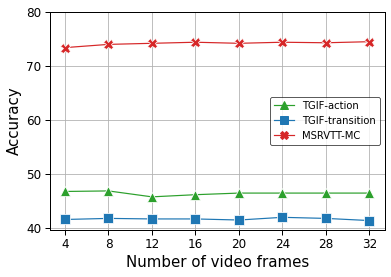}
         \caption{LaCLIP}
         \label{fig:laclip-videomc}
     \end{subfigure}
        \caption{Figure shows the effect of number of video frames on the zero-shot video multiple choice performance of CLIP and LaCLIP.}
        \label{fig:videomc_num_frames}
\end{figure}

%\subsection{Size of pre-training data matters for image-text models}
\subsection{Size of pre-training data matters for image-text models}
% Next, we analyse the effect of pre-training data size on the zero-shot performance of image-text models. We compare the image-text models like ALIP, SLIP pretrained on 15M data to CLIP, LaCLIP pretrained on 400M and OpenCLIP pretrained on 2B. 
We analyze the effect of the size of the pre-training data on the zero-shot performance of image-text models on the various video tasks.
We compare image-text models like ALIP and SLIP pretrained on 15M image-text pairs, CLIP and LaCLIP pretrained on 400M and OpenCLIP pretrained on 2B image-text pairs. On video AR and video MC as summarized in the Tables ~\ref{tab:act-ret-qa} and ~\ref{tab: video_mul_choice} respectively, we observe models trained with 400M data consistently significantly outperform the models trained with 15M data across all three evaluation datasets. These models also slightly outperform OpenCLIP trained on 400\% more training data. On the task of video retrieval, we note that larger pre-training data does have more impact on the zero-shot  performance gains on the MSVD dataset. In case of MSRVTT and DiDeMo, the models pretrained on 400M and 2B are comparable in performance. In all the evaluations, we observe that models trained on 15M image-text pairs performs 50\% worse than that of the models pretrained on 400M. On video QA, we observe that OpenCLIP performs remarkably well on MSRVTT compared to its counterparts which are trained on comparatively less data. However, on MSVD we see comparable performance between OpenCLIP, ALIP and SLIP trained on much lesser image-text pairs. When analyzing the performance on video captioning tasks in \ref{tab:video_cap}, we see BLIP2 trained with 129M image-text pairs outperforms OFA and Pix2Struct trained with lesser data (36.27M and 80M respectively). Overall we do observe a correlation between the amount of data used during pre-training and the zero-shot performance. However, the few outliers that we note do invite further analysis. 
%This probes us to evaluate the impact the pre-training objectives in the following section.

\subsection{Impact of frame count on zero-shot performance of image-text models}
In this section, we study the impact of video frame count on the zero-shot performance of image-text models. Figure ~\ref{fig:clip_num_frames} demonstrates the results of this study for CLIP model on video action recognition and video retrieval tasks. Our results indicate that the zero-shot performance increases with higher frame frequency and reaches an optimal point without further increase. We observe this effect for all the datasets of video action recognition and video retrieval. These results illustrate that frame count ranging from 12 to 20 offer the optimal zero-shot performance for most of the video tasks.

\section{Conclusion}
In this work, we presented a comprehensive analysis on the generalization capabilities of foundational image-text models on a broad range of video understanding tasks in a zero-shot setting. We evaluate nine foundational image-text models on five video tasks namely video action recognition, video retrieval, video question answering, video multiple choice and video captioning and also benchmark them against the state-of-the-art video-text models on these tasks. Our experiments demonstrate that image-text models accomplish competitive results to video-text models on video action recognition, video retrieval and video multiple choice. Moreover, the results also show that image-text models show reasonable performance on video captioning and perform worse on video question answering. We think that the future works can benefit from these findings in designing models for video understanding tasks.

%%%%%%%%%%%%%%%%%%%%%%%%%%%%%%%%%%%%%%%%%%%%%%%%%%%%%%%%%%%%
\bibliographystyle{ieee_fullname}
\bibliography{egbib,related}

\begin{thebibliography}{10}\itemsep=-1pt

\bibitem{agrawal2019nocaps}
Harsh Agrawal, Karan Desai, Yufei Wang, Xinlei Chen, Rishabh Jain, Mark Johnson, Dhruv Batra, Devi Parikh, Stefan Lee, and Peter Anderson.
\newblock Nocaps: Novel object captioning at scale.
\newblock In {\em Proceedings of the IEEE/CVF international conference on computer vision}, pages 8948--8957, 2019.

\bibitem{alamri2019audiovisual}
Huda Alamri, Vincent Cartillier, Abhishek Das, Jue Wang, Anoop Cherian, Irfan Essa, Dhruv Batra, Tim~K. Marks, Chiori Hori, Peter Anderson, Stefan Lee, and Devi Parikh.
\newblock Audio-visual scene-aware dialog.
\newblock In {\em Proceedings of the IEEE Conference on Computer Vision and Pattern Recognition}, 2019.

\bibitem{anne2017localizing}
Lisa Anne~Hendricks, Oliver Wang, Eli Shechtman, Josef Sivic, Trevor Darrell, and Bryan Russell.
\newblock Localizing moments in video with natural language.
\newblock In {\em Proceedings of the IEEE international conference on computer vision}, pages 5803--5812, 2017.

\bibitem{antol2015vqa}
Stanislaw Antol, Aishwarya Agrawal, Jiasen Lu, Margaret Mitchell, Dhruv Batra, C~Lawrence Zitnick, and Devi Parikh.
\newblock Vqa: Visual question answering.
\newblock In {\em Proceedings of the IEEE international conference on computer vision}, pages 2425--2433, 2015.

\bibitem{bain2021frozen}
Max Bain, Arsha Nagrani, G{\"u}l Varol, and Andrew Zisserman.
\newblock Frozen in time: A joint video and image encoder for end-to-end retrieval.
\newblock In {\em Proceedings of the IEEE/CVF International Conference on Computer Vision}, pages 1728--1738, 2021.

\bibitem{banerjee2005meteor}
Satanjeev Banerjee and Alon Lavie.
\newblock Meteor: An automatic metric for mt evaluation with improved correlation with human judgments.
\newblock In {\em Proceedings of the acl workshop on intrinsic and extrinsic evaluation measures for machine translation and/or summarization}, pages 65--72, 2005.

\bibitem{bossard2014food}
Lukas Bossard, Matthieu Guillaumin, and Luc Van~Gool.
\newblock Food-101--mining discriminative components with random forests.
\newblock In {\em Computer Vision--ECCV 2014: 13th European Conference, Zurich, Switzerland, September 6-12, 2014, Proceedings, Part VI 13}, pages 446--461. Springer, 2014.

\bibitem{changpinyo2021conceptual}
Soravit Changpinyo, Piyush Sharma, Nan Ding, and Radu Soricut.
\newblock Conceptual 12m: Pushing web-scale image-text pre-training to recognize long-tail visual concepts.
\newblock In {\em Proceedings of the IEEE/CVF Conference on Computer Vision and Pattern Recognition}, pages 3558--3568, 2021.

\bibitem{chen2011collecting}
David Chen and William~B Dolan.
\newblock Collecting highly parallel data for paraphrase evaluation.
\newblock In {\em Proceedings of the 49th annual meeting of the association for computational linguistics: human language technologies}, pages 190--200, 2011.

\bibitem{instructblip}
Wenliang Dai, Junnan Li, Dongxu Li, Anthony Meng~Huat Tiong, Junqi Zhao, Weisheng Wang, Boyang Li, Pascale Fung, and Steven Hoi.
\newblock Instructblip: Towards general-purpose vision-language models with instruction tuning, 2023.

\bibitem{das2017visual}
Abhishek Das, Satwik Kottur, Khushi Gupta, Avi Singh, Deshraj Yadav, Jos{\'e}~MF Moura, Devi Parikh, and Dhruv Batra.
\newblock Visual dialog.
\newblock In {\em Proceedings of the IEEE conference on computer vision and pattern recognition}, pages 326--335, 2017.

\bibitem{deng2009imagenet}
Jia Deng, Wei Dong, Richard Socher, Li-Jia Li, Kai Li, and Li Fei-Fei.
\newblock Imagenet: A large-scale hierarchical image database.
\newblock In {\em 2009 IEEE conference on computer vision and pattern recognition}, pages 248--255. Ieee, 2009.

\bibitem{dou2022empirical}
Zi-Yi Dou, Yichong Xu, Zhe Gan, Jianfeng Wang, Shuohang Wang, Lijuan Wang, Chenguang Zhu, Pengchuan Zhang, Lu Yuan, Nanyun Peng, et~al.
\newblock An empirical study of training end-to-end vision-and-language transformers.
\newblock In {\em Proceedings of the IEEE/CVF Conference on Computer Vision and Pattern Recognition}, pages 18166--18176, 2022.

\bibitem{fan2023improving}
Lijie Fan, Dilip Krishnan, Phillip Isola, Dina Katabi, and Yonglong Tian.
\newblock Improving clip training with language rewrites.
\newblock {\em arXiv preprint arXiv:2305.20088}, 2023.

\bibitem{fei2004learning}
Li Fei-Fei, Rob Fergus, and Pietro Perona.
\newblock Learning generative visual models from few training examples: An incremental bayesian approach tested on 101 object categories.
\newblock In {\em 2004 conference on computer vision and pattern recognition workshop}, pages 178--178. IEEE, 2004.

\bibitem{fu2021violet}
Tsu-Jui Fu, Linjie Li, Zhe Gan, Kevin Lin, William~Yang Wang, Lijuan Wang, and Zicheng Liu.
\newblock Violet: End-to-end video-language transformers with masked visual-token modeling.
\newblock 2021.

\bibitem{fu2023empirical}
Tsu-Jui Fu, Linjie Li, Zhe Gan, Kevin Lin, William~Yang Wang, Lijuan Wang, and Zicheng Liu.
\newblock An empirical study of end-to-end video-language transformers with masked visual modeling.
\newblock In {\em Proceedings of the IEEE/CVF Conference on Computer Vision and Pattern Recognition}, pages 22898--22909, 2023.

\bibitem{ilharco_gabriel_2021_5143773}
Gabriel Ilharco, Mitchell Wortsman, Ross Wightman, Cade Gordon, Nicholas Carlini, Rohan Taori, Achal Dave, Vaishaal Shankar, Hongseok Namkoong, John Miller, Hannaneh Hajishirzi, Ali Farhadi, and Ludwig Schmidt.
\newblock Openclip, July 2021.
\newblock If you use this software, please cite it as below.

\bibitem{jang2017tgif}
Yunseok Jang, Yale Song, Youngjae Yu, Youngjin Kim, and Gunhee Kim.
\newblock Tgif-qa: Toward spatio-temporal reasoning in visual question answering.
\newblock In {\em Proceedings of the IEEE conference on computer vision and pattern recognition}, pages 2758--2766, 2017.

\bibitem{krizhevsky2009learning}
Alex Krizhevsky, Geoffrey Hinton, et~al.
\newblock Learning multiple layers of features from tiny images.
\newblock 2009.

\bibitem{kuehne2011hmdb}
Hildegard Kuehne, Hueihan Jhuang, Est{\'\i}baliz Garrote, Tomaso Poggio, and Thomas Serre.
\newblock Hmdb: a large video database for human motion recognition.
\newblock In {\em 2011 International conference on computer vision}, pages 2556--2563. IEEE, 2011.

\bibitem{lee2023pix2struct}
Kenton Lee, Mandar Joshi, Iulia~Raluca Turc, Hexiang Hu, Fangyu Liu, Julian~Martin Eisenschlos, Urvashi Khandelwal, Peter Shaw, Ming-Wei Chang, and Kristina Toutanova.
\newblock Pix2struct: Screenshot parsing as pretraining for visual language understanding.
\newblock In {\em International Conference on Machine Learning}, pages 18893--18912. PMLR, 2023.

\bibitem{lei2020tvr}
Jie Lei, Licheng Yu, Tamara~L Berg, and Mohit Bansal.
\newblock Tvr: A large-scale dataset for video-subtitle moment retrieval.
\newblock In {\em Computer Vision--ECCV 2020: 16th European Conference, Glasgow, UK, August 23--28, 2020, Proceedings, Part XXI 16}, pages 447--463. Springer, 2020.

\bibitem{li2022align}
Dongxu Li, Junnan Li, Hongdong Li, Juan~Carlos Niebles, and Steven~CH Hoi.
\newblock Align and prompt: Video-and-language pre-training with entity prompts.
\newblock In {\em Proceedings of the IEEE/CVF Conference on Computer Vision and Pattern Recognition}, pages 4953--4963, 2022.

\bibitem{li2023blip}
Junnan Li, Dongxu Li, Silvio Savarese, and Steven Hoi.
\newblock Blip-2: Bootstrapping language-image pre-training with frozen image encoders and large language models.
\newblock 2023.

\bibitem{li2022blip}
Junnan Li, Dongxu Li, Caiming Xiong, and Steven~CH Hoi.
\newblock Blip: Bootstrapping language-image pre-training for unified vision-language understanding and generation.
\newblock 2022.

\bibitem{li2021align}
Junnan Li, Ramprasaath Selvaraju, Akhilesh Gotmare, Shafiq Joty, Caiming Xiong, and Steven Chu~Hong Hoi.
\newblock Align before fuse: Vision and language representation learning with momentum distillation.
\newblock {\em Advances in neural information processing systems}, 34:9694--9705, 2021.

\bibitem{li2020hero}
Linjie Li, Yen-Chun Chen, Yu Cheng, Zhe Gan, Licheng Yu, and Jingjing Liu.
\newblock Hero: Hierarchical encoder for video+ language omni-representation pre-training.
\newblock In {\em Proceedings of the 2020 Conference on Empirical Methods in Natural Language Processing (EMNLP)}, pages 2046--2065, 2020.

\bibitem{li2023lavender}
Linjie Li, Zhe Gan, Kevin Lin, Chung-Ching Lin, Zicheng Liu, Ce Liu, and Lijuan Wang.
\newblock Lavender: Unifying video-language understanding as masked language modeling.
\newblock In {\em Proceedings of the IEEE/CVF Conference on Computer Vision and Pattern Recognition}, pages 23119--23129, 2023.

\bibitem{li2019visualbert}
Liunian~Harold Li, Mark Yatskar, Da Yin, Cho-Jui Hsieh, and Kai-Wei Chang.
\newblock Visualbert: Asimple and performant baseline for vision and language.
\newblock {\em arXiv preprint arXiv:1908.03557}, 2019.

\bibitem{li2020unimo}
Wei Li, Can Gao, Guocheng Niu, Xinyan Xiao, Hao Liu, Jiachen Liu, Hua Wu, and Haifeng Wang.
\newblock Unimo: Towards unified-modal understanding and generation via cross-modal contrastive learning.
\newblock {\em arXiv preprint arXiv:2012.15409}, 2020.

\bibitem{li2020oscar}
Xiujun Li, Xi Yin, Chunyuan Li, Pengchuan Zhang, Xiaowei Hu, Lei Zhang, Lijuan Wang, Houdong Hu, Li Dong, Furu Wei, et~al.
\newblock Oscar: Object-semantics aligned pre-training for vision-language tasks.
\newblock In {\em Computer Vision--ECCV 2020: 16th European Conference, Glasgow, UK, August 23--28, 2020, Proceedings, Part XXX 16}, pages 121--137. Springer, 2020.

\bibitem{li2021supervision}
Yangguang Li, Feng Liang, Lichen Zhao, Yufeng Cui, Wanli Ouyang, Jing Shao, Fengwei Yu, and Junjie Yan.
\newblock Supervision exists everywhere: A data efficient contrastive language-image pre-training paradigm.
\newblock In {\em International Conference on Learning Representations}, 2021.

\bibitem{lin2014microsoft}
Tsung-Yi Lin, Michael Maire, Serge Belongie, James Hays, Pietro Perona, Deva Ramanan, Piotr Doll{\'a}r, and C~Lawrence Zitnick.
\newblock Microsoft coco: Common objects in context.
\newblock In {\em Computer Vision--ECCV 2014: 13th European Conference, Zurich, Switzerland, September 6-12, 2014, Proceedings, Part V 13}, pages 740--755. Springer, 2014.

\bibitem{lin2023match}
Wei Lin, Leonid Karlinsky, Nina Shvetsova, Horst Possegger, Mateusz Kozinski, Rameswar Panda, Rogerio Feris, Hilde Kuehne, and Horst Bischof.
\newblock Match, expand and improve: Unsupervised finetuning for zero-shot action recognition with language knowledge.
\newblock {\em arXiv preprint arXiv:2303.08914}, 2023.

\bibitem{miech2019howto100m}
Antoine Miech, Dimitri Zhukov, Jean-Baptiste Alayrac, Makarand Tapaswi, Ivan Laptev, and Josef Sivic.
\newblock Howto100m: Learning a text-video embedding by watching hundred million narrated video clips.
\newblock In {\em Proceedings of the IEEE/CVF International Conference on Computer Vision}, pages 2630--2640, 2019.

\bibitem{mu2022slip}
Norman Mu, Alexander Kirillov, David Wagner, and Saining Xie.
\newblock Slip: Self-supervision meets language-image pre-training.
\newblock In {\em European Conference on Computer Vision}, pages 529--544. Springer, 2022.

\bibitem{papineni2002bleu}
Kishore Papineni, Salim Roukos, Todd Ward, and Wei-Jing Zhu.
\newblock Bleu: a method for automatic evaluation of machine translation.
\newblock In {\em Proceedings of the 40th annual meeting of the Association for Computational Linguistics}, pages 311--318, 2002.

\bibitem{radford2021learning}
Alec Radford, Jong~Wook Kim, Chris Hallacy, Aditya Ramesh, Gabriel Goh, Sandhini Agarwal, Girish Sastry, Amanda Askell, Pamela Mishkin, Jack Clark, et~al.
\newblock Learning transferable visual models from natural language supervision.
\newblock In {\em International Conference on Machine Learning}, pages 8748--8763. PMLR, 2021.

\bibitem{sharma2018conceptual}
Piyush Sharma, Nan Ding, Sebastian Goodman, and Radu Soricut.
\newblock Conceptual captions: A cleaned, hypernymed, image alt-text dataset for automatic image captioning.
\newblock In {\em Proceedings of the 56th Annual Meeting of the Association for Computational Linguistics (Volume 1: Long Papers)}, pages 2556--2565, 2018.

\bibitem{smaira2020short}
Lucas Smaira, Jo{\~a}o Carreira, Eric Noland, Ellen Clancy, Amy Wu, and Andrew Zisserman.
\newblock A short note on the kinetics-700-2020 human action dataset.
\newblock {\em arXiv preprint arXiv:2010.10864}, 2020.

\bibitem{soomro2012ucf101}
Khurram Soomro, Amir~Roshan Zamir, and Mubarak Shah.
\newblock Ucf101: A dataset of 101 human actions classes from videos in the wild.
\newblock {\em arXiv preprint arXiv:1212.0402}, 2012.

\bibitem{vedantam2015cider}
Ramakrishna Vedantam, C Lawrence~Zitnick, and Devi Parikh.
\newblock Cider: Consensus-based image description evaluation.
\newblock In {\em Proceedings of the IEEE conference on computer vision and pattern recognition}, pages 4566--4575, 2015.

\bibitem{wang2022ofa}
Peng Wang, An Yang, Rui Men, Junyang Lin, Shuai Bai, Zhikang Li, Jianxin Ma, Chang Zhou, Jingren Zhou, and Hongxia Yang.
\newblock Ofa: Unifying architectures, tasks, and modalities through a simple sequence-to-sequence learning framework.
\newblock In {\em International Conference on Machine Learning}, pages 23318--23340. PMLR, 2022.

\bibitem{xu2017video}
Dejing Xu, Zhou Zhao, Jun Xiao, Fei Wu, Hanwang Zhang, Xiangnan He, and Yueting Zhuang.
\newblock Video question answering via gradually refined attention over appearance and motion.
\newblock In {\em Proceedings of the 25th ACM international conference on Multimedia}, pages 1645--1653, 2017.

\bibitem{xu2016msr}
Jun Xu, Tao Mei, Ting Yao, and Yong Rui.
\newblock Msr-vtt: A large video description dataset for bridging video and language.
\newblock In {\em Proceedings of the IEEE conference on computer vision and pattern recognition}, pages 5288--5296, 2016.

\bibitem{yang2023vid2seq}
Antoine Yang, Arsha Nagrani, Paul~Hongsuck Seo, Antoine Miech, Jordi Pont-Tuset, Ivan Laptev, Josef Sivic, and Cordelia Schmid.
\newblock Vid2seq: Large-scale pretraining of a visual language model for dense video captioning.
\newblock In {\em Proceedings of the IEEE/CVF Conference on Computer Vision and Pattern Recognition}, pages 10714--10726, 2023.

\bibitem{Yang_2022_CVPR}
Jianwei Yang, Chunyuan Li, Pengchuan Zhang, Bin Xiao, Ce Liu, Lu Yuan, and Jianfeng Gao.
\newblock Unified contrastive learning in image-text-label space.
\newblock In {\em Proceedings of the IEEE/CVF Conference on Computer Vision and Pattern Recognition (CVPR)}, pages 19163--19173, June 2022.

\bibitem{yang2023alip}
Kaicheng Yang, Jiankang Deng, Xiang An, Jiawei Li, Ziyong Feng, Jia Guo, Jing Yang, and Tongliang Liu.
\newblock Alip: Adaptive language-image pre-training with synthetic caption.
\newblock {\em arXiv preprint arXiv:2308.08428}, 2023.

\bibitem{young2014image}
Peter Young, Alice Lai, Micah Hodosh, and Julia Hockenmaier.
\newblock From image descriptions to visual denotations: New similarity metrics for semantic inference over event descriptions.
\newblock {\em Transactions of the Association for Computational Linguistics}, 2:67--78, 2014.

\bibitem{yu2018joint}
Youngjae Yu, Jongseok Kim, and Gunhee Kim.
\newblock A joint sequence fusion model for video question answering and retrieval.
\newblock In {\em Proceedings of the European Conference on Computer Vision (ECCV)}, pages 471--487, 2018.

\bibitem{yuan2021florence}
Lu Yuan, Dongdong Chen, Yi-Ling Chen, Noel Codella, Xiyang Dai, Jianfeng Gao, Houdong Hu, Xuedong Huang, Boxin Li, Chunyuan Li, et~al.
\newblock Florence: A new foundation model for computer vision.
\newblock {\em arXiv preprint arXiv:2111.11432}, 2021.

\bibitem{zellers2022merlot}
Rowan Zellers, Jiasen Lu, Ximing Lu, Youngjae Yu, Yanpeng Zhao, Mohammadreza Salehi, Aditya Kusupati, Jack Hessel, Ali Farhadi, and Yejin Choi.
\newblock Merlot reserve: Neural script knowledge through vision and language and sound.
\newblock In {\em Proceedings of the IEEE/CVF Conference on Computer Vision and Pattern Recognition}, pages 16375--16387, 2022.

\bibitem{ZhXuCoAAAI18}
Luowei Zhou, Chenliang Xu, and Jason~J Corso.
\newblock Towards automatic learning of procedures from web instructional videos.
\newblock In {\em AAAI Conference on Artificial Intelligence}, pages 7590--7598, 2018.

\end{thebibliography}
\newpage

\appendix

\end{document}